\documentclass{article}
\usepackage{iclr2026_conference,times}

\usepackage{microtype}
\usepackage{graphicx}
\usepackage{subcaption}
\usepackage{booktabs} 

\usepackage{hyperref}
\usepackage{xurl}

\usepackage{amsmath}
\usepackage{amssymb}
\usepackage{mathtools}
\usepackage{amsthm}

\usepackage{csquotes}

\usepackage[capitalize,noabbrev]{cleveref}

\theoremstyle{plain}

\theoremstyle{definition}

\theoremstyle{remark}

\title{\fontsize{16pt}{16pt}\selectfont The implicated scientist: on the role of AI researchers in the development of weapons systems\thanks{Presented as an oral talk and a poster at the \href{https://aiforpeaceworkshop.github.io/}{AI for Peace workshop} at ICLR 2026.}}

\author{Alexandra Volokhova, Alex Hernandez-Garcia\\
Université de Montréal, Mila -- Quebec AI Institute\\
Montreal, QC, Canada \\
\texttt{\{alexandra.volokhova, hernanga\}@mila.quebec} \\
}

\iclrfinalcopy 
\begin{document}

\maketitle

\begin{abstract}
  Artificial intelligence (AI) technologies are increasingly used in modern weapons systems. Notably, these systems have recently been involved in mass killings and destruction at scale. Furthermore, there is currently a strong interest and competition among powerful players to accelerate the proliferation of weapons with automated or AI-based components, a phenomenon known as ``AI arms race''. This competition poses a risk of causing even more deaths and devastation in the future, as well as increased power and wealth inequality. In this work, we aim to shed light on the role of AI researchers as \textit{implicated subjects} in the harms caused by weapons enabled by AI technologies. We investigate and discuss the specifics of this implication and explore ways to transfigure this position of implication into one of differentiated, long-distance solidarity with the victims of technologically fortified injustices.
\end{abstract}

\section{Introduction}
\label{sec:intro}

The work of scientists, journalists, and activists has recently highlighted the profound influence of machine learning and data science---often refereed to as  artificial intelligence (AI)---on modern technologies of warfare and surveillance. For example, \citet{kalluri2023surveillance} have recently revealed the \textit{surveillance AI pipeline}, through which AI research has enabled the development of technology for mass surveillance \citep{amnesty2023}; \citet{marijan2023autonomousweapons} have warned about the risks of the ongoing weaponization of AI, with autonomous weapons and deepfakes; and \citet{simmons2024ai} have discussed the impacts on geopolitical instability and AI research itself.
In extreme cases, it has been reported that that some of these technologies have been used for mass killings and destruction \citep{abraham2024lavender, bonea2025algorithmsofwar}. 

Until recently, the involvement of technology companies in military applications was kept as confidential as possible because it elicited negative responses \citep{marijan2025bigtechmilitary}. However, leading tech companies are breaking this taboo and are expanding and announcing their collaborations with military agencies and weapons manufacturers \citep{biddle2024,vynck2024,wiggers2024,tiku2025}, notwithstanding their employees protests against such military contracts \citep{perrigo2024,ntfa2025}. These changes are happening in a context of increased military spending \citep{nato2025fivepercent}, characterised by large investments in AI-enabled warfare technology, which some authors have referred to as an ``AI arms race'' \citep{maaser2022big, gonzalez2024bigtech}.

In this context, AI researchers in both academia and the private sector are part of a deeply woven network which fuels the competition for technological domination and puts them in a position of relative power. This power brings privilege and benefits to the researchers, such as funding opportunities \citep{larson2024}, high salaries \citep{harroch2026aicompensation}, travelling opportunities and social status. However, the fact that the work of AI researchers is becoming closely associated with the technological arms race and related harms, is also raising ethical concerns among many scientists in the AI community and beyond \citep{marijan2023autonomousweapons, simmons2024ai}.

Here, we propose to study the position of AI researchers in the context of an AI arms race using the figure of the \textit{implicated subject} and the related notion of \textit{implication} into the socioeconomic structures perpetuating injustice and oppression \citep{rothberg2019implicated}. This vocabulary, uncommon within the technical AI community, allows us to critically examine personal and collective contributions to and benefits from the structures creating AI-enabled weapons systems and their associated harms\footnote{This study emphasizes the connection between AI research and weapons systems---technologies which enable or cause devastating harms of killing and destruction in the most direct way. The technologies considered include (semi)autonomous weapons, automated targeting systems, and to a lesser extent surveillance systems.}. Our examination brings an insight for possible strategies of individual and collective actions directed towards structural transformations and, ultimately, towards the collective liberation from oppressive and violent systems.

\section{Background: The implicated subject}
\label{sec:background}
 
The present study is largely inspired by the notion of \textit{implicated subject}, proposed by \citet{rothberg2019implicated}:

\begin{displayquote}
Implicated subjects occupy positions aligned with power and privilege without being themselves direct agents of harm; they contribute to, inhabit, inherit or benefit from regimes of domination but do not originate or control such regimes. [...] Although indirect or belated, their actions and inactions help produce and reproduce the positions of victims and perpetrators. 
\end{displayquote}

Focusing on the figure of the implicated subject, we direct our attention towards the building blocks of the oppressive system, recognise ourselves as a part of it and explore potential pathways towards structural change. The implicated position is not defining of one's identity, but it instead describes their socioeconomic position with respect to an event of injustice or harm. Therefore, implication does not suggest a sentiment of blame or guilt, but invites an attentive reflection on individual contributions to the system of oppression and available leverages for its transformation. 

As \citet{rothberg2019implicated} suggests, implicated subjects can transfigure their position by engaging with the work of solidarity, that is making tangible efforts for resisting the oppression and supporting its victims. In these efforts, Rothberg emphasizes the importance of differentiated, long-distance solidarity, which acknowledges the differences and distances between the positions of various subjects involved, yet recognises their social connectedness and ``attempts to incite a change in and from the place'' of one's own position. Importantly, these attempts are most effective when performed collectively and essentially start with confronting the ``socially sanctioned denial and ignorance'' of the implication.

Importantly, multiple systems of oppression co-exist simultaneously with each other and with diverse legacies of injustice. Therefore, most people find themselves implicated in different oppressive systems, past or present, where their positions are sometimes closer to victims, and other times to perpetrators. Rothberg calls this phenomenon ``complex implication''. Moreover, the position of implication is not permanent, as it changes dynamically together with one's life events and transitions. The resulting complexity and dynamism are important characteristics of the implication position which have to be considered when we aim to transfigure our implication into solidarity.  


\section{Implication of AI researchers in the military}\label{sec:implication}



  


\paragraph{Historical perspective} The implication of AI research in the military can be traced back to the historical roots of AI as a research field. As described by \citet{katz2020artificial}, since the 1960s, shortly after defining AI as a new research field, practitioners have been steadily receiving significant funding from the United States Department of Defense. Since the inception of the field, the research aspirations of making a computer perform ``intelligent tasks'' have developed under the influence of military patrons and their imaginaries of automated battlefields. \citet{katz2020artificial} suggests that this close relationship deeply affected the core aspects of the AI research methodology, including the very notion of ``intelligence'' operationalised in AI research.
Under this notion, humans are perceived as ``information processing systems'' characterized as ``intelligent'' when solving abstract puzzles. This operational perspective easily accommodates a wide range of military tasks, for example object and speech recognition for surveillance.
Furthermore, AI methodology is largely focused on accumulating knowledge, increasing understanding and control, similarly to cybernetics, which makes both fields very suitable for the military applications serving authoritarian goals \citep{katz2020artificial, ekbia2023killing}.

Later, in the 1980s, the concept of ``dual use'' was developed in the U.S. as a strategy for managing and expanding military innovation, including in the realm of digital technologies \citep{maaser2022big}. In this innovation process, the duality is not limited to the end use of a technology but instead arises from a process of the cooperative knowledge creation and dissemination between civilian and military agents \citep{meunier2019construction}. Notably, the specifics of this cooperation largely determine the dual nature of the technology \citep{te2003civilian}. In this context, the AI research field and military development have co-evolved in a close alliance sharing expertise, talent and knowledge.

\paragraph{Military funding of ``basic research''} The entanglement of AI research and the military in the U.S. context has been recently analysed by \citet{widder2024basicresearch}. The authors of this paper examined a corpus of grant solicitations between 2007 and 2023 from the U.S. Department of Defense (DoD). One of the conclusions of the analysis reinforces the notion that military technology and AI research co-evolve in a synergistic way, which the paper describes as ``mutual enlistment''. Another important conclusion concerns the classic distinction between ``basic'' and ``applied'' research. The common distinction holds that ``basic research is considered to be conducted without ostensible intentions for particular end uses''. However, \citet{widder2024basicresearch} critically examine this distinction and conclude, from their analysis of DoD grant solicitations, that the category of basic research serves the purpose of creating a moral shelter for scientists, while it nonetheless enlists them in a military agenda. For example, the authors discuss indicators that proposals for funding programs categorised as basic research have higher chances of success if they are aligned with DoD objectives. This is often stated explicitly, for instance, by setting as criteria the establishment of ``long-term relationships between university researchers and the DoD'', or training students ``for the defense workforce'' \citep{widder2024basicresearch}.

These conclusions become all the more relevant in light of the recent increase in AI spending by the U.S. DoD: From 2022 to 2023, the potential value of AI-related spending by the DoD rose from \$269 million U.S. dollars to \$4,323 million (16-fold increase), which comprises 95~\% of all AI-related federal funding \citep{larson2024}. Beyond the U.S., the European Union has also recently reached record levels of military spending \citep{eu2025defense}, and (most) NATO members have agreed to increase their military spending to 5~\% of their GDP by 2035 \citep{nato2025fivepercent}. Crucially, at least part of the extended military budget is expected to be dedicated to fund academic research. For example, Canada plans to increase ``defence-related research and development'' by 85~\%, and AI is highlighted as a key sector \citep{canada2026defence, borealis2025}. The European Union and its member states are also increasing their military research budgets, and developing out specific AI strategies \citep{clapp2025defenceAI}. A consequence for academics is that their research funding opportunities will be increasingly connected to the military.

\paragraph{Industry involvement} The development of AI research, and its translation into military applications, cannot be fully understood without the role of the industry, and in particular big tech. On the one hand, big tech companies have become influential actors in AI research, contributing around a quarter of the publications at AI academic conferences, among other indicators \citep{abdalla2021bigtechbigtobacco, ahmed2023industry, hg2025iai}. This permeation in the field has allowed big tech to largely influence the agenda of AI research. On the other hand, the involvement of big tech---and the AI industry more generally---in militarism has also increased significantly \citep{maaser2022big}. Beyond the collaboration with the U.S. DoD \citep{coveri2025blurring}, Google, Microsoft and Amazon have been reported by journalistic investigation to tightly collaborate with the Israeli military \citep{fatafta2025artificial, abraham2024order, davies2025revealed, ohchr2025from}---since late 2023, the State of Israel has an open proceeding at the International Court of Justice concerning alleged violations in the Gaza Strip of the Convention on the Prevention and Punishment of the Crime of Genocide \citep{icj2024genocide}. Furthermore, these big tech companies have also close ties with both the major weapons manufacturers (known as \textit{primes}) and the emerging military startups (\textit{neoprimes}). For instance, Google and Lockheed Martin, one of the largest weapons manufacturers, recently announced a collaboration involving generative AI \citep{lockheed2025google}. Microsoft and Meta have also signed agreements with Lockheed Martin and with neoprimes such as Palantir and Anduril \citep{microsoft2025lockheed, gonzalez2024bigtech}. Finally, there are also cases where weapons manufacturers are directly involved in the AI research community, as is the case of Helsing, a German-based startup that produces AI-enabled killer drones. The chief scientist of Helsing is a former co-managing director at Facebook AI Research and researchers in the company have co-authored papers presented at major AI conferences \citep{franceschi2024explaining}.

\paragraph{Conclusion} When all these insights are jointly considered, it is possible to trace the implication of AI researchers---in the sense of the concept described in \cref{sec:background}---in the proliferation of AI-enabled weapons and their devastating consequences: The AI field has evolved since its inception thanks to strong ties with the military agencies; these military agencies manage to influence the AI research agenda through their funding system, while they also establish strong ties with big tech and weapons manufacturers; big tech is largely implicated in both the development of AI research as well as military applications, through collaborations with state armies and weapons manufacturers. The most tangible consequence of this mesh of relationships is that AI technologies that have been recently developed within the research ecosystem are today essential components of modern autonomous weapons and other military systems. Under the framework of implication \citep{rothberg2019implicated}, AI researchers are situated as contributors to the development of these technologies and as recipients of associated personal benefits, that is as implicated subjects. Rather than to assign blame or guilt, the position of implication can be used to envision effective strategies of transformation, which we discuss in \cref{sec:resist}.




\section{How to resist: transfiguring implication into solidarity}
\label{sec:resist}


As we discussed in \cref{sec:background}, implicated subjects are embedded in the oppressive system. By their actions and inaction, they collectively support the system and enable its harms. Therefore, they have a potential to transform the system---by acting differently individually and collectively. In this section we suggest directions for such transformative efforts in the context of AI researchers' implication in the weapons systems development.

\paragraph{Challenge epistemic assumptions} The most fundamental transformation is needed at the core foundations of the field. Instead of assuming the operational ``universal'' model of intelligence and the world, feminist and decolonial scholars suggest embracing diversity of forms of the situated knowledge, which is ``produced by specific people in specific circumstances---cultural, historical, and geographic'' \citep{d2023data}. They reject the separation of ``feeling from being, knowing, doing or living'' and value the knowledge that comes from the lived experience \citep{manyfesto}. This paradigm shifts the direction of thought from the abstract puzzle solving in the service of ultimate control towards centring diverse human experiences, acceptance of the pluralism and the unknown. These ideas are alien to the military world and create a foundation for developing radically different technologies, which empower people and communities instead of the empires. The decolonial and feminist perspectives lead to research methodologies prioritising data sovereignty, participatory design, social relations, ethical labour, and the development of situated narrowly-scoped technologies \citep{bender2025aicon, carroll2023care}.

\paragraph{Develop critical approaches} Fundamental and structural transformations happen in non-linear ways \citep{duncan2016change}. We suggest treating them as processes rather than the end-goals and search for the ways to tangibly contribute to them. As AI researchers, we shall start with combating the denial of our implication---within ourselves, our institutions, and our research community. Education in critical theory and thinking plays crucial role in this process and requires conscious effort, as STEM curriculums rarely cover these subjects. We recommend, when possible, creating spaces for critical discussions at universities, workplaces, and conferences, as this contributes to both rasing awareness and strengthening the community for collective action. Furthermore, academic studies scrutinising power and values driving AI research provide valuable insight into the effective strategies and targets for transformative action. 

\paragraph{Refuse military collaborations} Certainly, avoiding military funding and collaborations helps liberate one's research from the direct influence of military incentives. As mentioned in \cref{sec:implication}, even the ``basic'' academic research funded by a military agency contributes to the dual-use innovation system. This increases the proximity between the academia and the military, normalising this relationship, facilitating knowledge and talent exchange and therefore, deepening the contribution of academic researchers to weapons development. When possible, we suggest to pursue more ethical funding sources, such as government funding for civilian science, professional societies grants, trustworthy philanthropic foundations, industry partnerships with an ethical mission, and citizen science crowdfunding. When it is not possible to avoid military funding, we suggest to put increased effort into the critical evaluation of one's own research, its contribution to the military agendas and possible ways to sabotage them through collective direct action. 

\paragraph{Build coalitions} Beyond the frame of the conventional scholarly activities, we highlight the importance of coalition building, unionisation, direct action, and engagement with relevant organizations and political movements. For example, Science for the People \citep{sftp2025} is a decentralized organisation committed to radical science, which was founded in the U.S. at the height of the anti–war movement, and has since expanded to have chapters around the world. SftP publishes a magazine with critical articles about science, technology, and society, which have contributed to the anti-military discourse on AI and computing \citep{weizenbaum1985computers, ekbia2023killing}. Other relevant movements are the Stop Killer Robots campaign, which advocates for international regulations of the autonomy in weapons systems \citep{skr}, and Reaching Critical Will, which upholds disarmament and arms control across many weapon systems from feminist, queer, and anti-racist perspectives \citep{rcw}.

\paragraph{Direct action} Finally, the actions of tech workers, such as open letters, petitions, sit-ins, strikes and whistleblowing \citep{twpb}, are critically valuable resistance strategies. A prominent example is the campaign of Google workers against Project Maven, a contract between Google and the U.S. DoD for developing computer vision technologies for miliary drones \citep{pbs2018maven}. In this effort, workers pressured the company not to renew the contract with the DoD, which slowed down Google's rapprochement with the military. Several years later, Google and Amazon signed cloud computing contracts with the Israeli government and military, and their workers responded with the No Tech for Apartheid campaign \citep{ntfa2025, koren2022google}. Activists of the campaign organise sit-ins in the offices, expose companies' military contracts and call for an AI arms embargo \citep{ntfa2025tellgoogle}. Similarly, Microsoft has also been found to tightly collaborate with the Israeli's army cyber warfare unit on the mass storage of surveillance data about Palestinians \citep{abraham2025microsoft8200}. The collective efforts of workers through the No Azure for Apartheid campaign helped force Microsoft reduce its ties with the Israeli military \citep{nafa2024, davies2025microsoft}.

\paragraph{Embrace complexity} As mentioned in \cref{sec:background}, everyone's position of implication is uniquely complex as we exist within the interlocking histories and actualities of oppression. Therefore, when we consider transformative actions, it is essential to distinguish between different levels of agency, autonomy, and institutional power. For example, early-career researchers often operate under significant structural constraints, whereas senior researchers and tenured professors may have substantial influence over funding decisions, research agendas, and partnerships. Furthermore, there exist differences within the groups of the same career level. For instance, international students might bare additional risks when engaging in direct action than local students. Moreover, these contexts constantly evolve together with one's career and life situation. All these aspects highlight that collective resistance cannot be distilled to a list of unified recommendations uniformly applicable to everyone. Instead, each person has to put effort into a critical evaluation of their own context and available leverages for contributing to the systemic transformation.

We encourage AI researchers to strive for upholding humanistic values in every professional and personal choice of their life. We find value-based honest critical thinking to be the most helpful in guiding personal decisions for specific solidarity strategies.


\section{Conclusion}

In this paper, we have examined how AI research is implicated in the harms caused by AI-enhanced weapons systems. In spite of the fact that individual researchers are usually not in a position to take political decisions about military attacks and conflicts, they still have tangible leverages for resisting the harmful uses of their field's scientific advances. In this work, we have argued that the recognition and the active transformation of one's position of implication through the work of solidarity is crucial for resisting the militarized oppressive system. We believe this can bring structural transformation, when adopted widely. Further, we have provided suggestions for individual and collective resistance strategies and encouraged the reader to reflect on their own areas of influence to find possible steps they can take for transfiguring their own position of implication into one of solidarity.

\section*{Positionality statement}

This work presents a critical perspective informed by the authors' belief that every life is invaluable and that the act of killing a human is an act of a profound, unrepairable harm. In addition, we believe in the human intrinsic ability and desire to uphold humanistic values and to thrive in cooperation with others and the world.

As white, middle-class AI researchers at a Canadian University, we acknowledge our position in relation to this study on implication of AI researchers in the development of weapons systems. We acknowledge that our privileged position means that we may have blind spots regarding the realities of researchers coming from different background. We also cannot fully understand the experiences of communities directly affected by AI militarisation. In addition, our professional pathways have been mostly technical and therefore we lack the depth that the social sciences perspective can offer on this topic. We have attempted to minimise the limitations of our position through solidarity efforts and ongoing reflection throughout the research process.

\section*{Acknowledgements}

We would like to express our sincere acknowledgement to Ezekiel Williams for his significant contribution to the earlier stages of this project. We are also grateful to Matt Kusner for his support and participation in discussions. Alexandra Volokhova is grateful to her supervisor, Prof. Yoshua Bengio for supporting her work on this project and other related initiatives. Alex Hernandez-Garcia acknowledges funding from IVADO and the Canada First Research Excellence Fund. This article also owes credit to the participants and co-organisers of Mila's \href{https://criticalscience.github.io/}{Critical Science reading group}, and earlier, the reading group Against Military AI, and the workshop \href{https://www.harms-risks-ai-military.org}{Harms and Risks of AI in the Military}. These discussions have provided us much inspiration and they laid the foundations of the research that led to this article. Finally, we acknowledge that we are only humbly standing on the shoulders of many brave activists and scholars who have campaigned and written against militarism.



\bibliography{references}
\bibliographystyle{iclr2026_conference}




\end{document}